\documentclass[letterpaper, 10 pt, conference]{ieeeconf}  

\IEEEoverridecommandlockouts                              

\usepackage{geometry}
\usepackage{algorithm} 
\usepackage{program} 
\usepackage{algorithmic}
\usepackage{listings}
\usepackage{graphicx}
\usepackage{wrapfig}
\usepackage{comment}
\usepackage{hyperref}
\usepackage{listings}
\usepackage{caption}
\usepackage{subcaption}
\usepackage{xcolor}
\usepackage{placeins} 

\renewcommand{\footnotesize}{\scriptsize}

\usepackage{caption,setspace}
\DeclareCaptionFormat{listing}{\rule{\dimexpr0.9\columnwidth+17pt\relax}{0.4pt}\par\vskip1pt#1#2#3}
\newcounter{code}



\definecolor{redalias}{HTML}{CA1236}

\title{
Can ROS be used securely in industry? \\
Red teaming ROS-Industrial
}

\author{Víctor Mayoral-Vilches$^{1,}$$^{2}$, Martin Pinzger$^{1}$, Stefan Rass$^{1}$, \\
    Bernhard Dieber$^{3}$ and Endika Gil-Uriarte$^{2}$ 
\thanks{$^{1}$Víctor Mayoral-Vilches and Stefan Rass are with the System Security (SYSSEC) group. Martin Pinzger is with the Software Engineering group. Both from Universität Klagenfurt, Austria.
        {\tt\footnotesize v1mayoralv@edu.aau.at}}%
\thanks{$^{2}$Víctor Mayoral-Vilches and Endika Gil-Uriarte are with Alias Robotics,  Spain}%
\thanks{$^{3}$Bernhard Dieber is with the Institute for Robotics and Mechatronics, Joanneum Research, Austria}%
}

\begin{document}

\maketitle
\pagestyle{plain}

\begin{abstract}

With its growing use in industry, ROS is rapidly becoming a standard in robotics. While developments in ROS 2 show promise, the slow adoption cycles in industry will push widespread ROS 2 industrial adoption years from now. ROS will prevail in the meantime which raises the question: can ROS be used securely for industrial use cases even though its origins didn't consider it? The present study analyzes this question experimentally by performing a targeted offensive security exercise in a synthetic industrial use case involving ROS-Industrial and ROS packages. 
Our exercise results in four groups of attacks which manage to compromise the ROS computational graph, and all except one take control of most robotic endpoints at desire. To the best of our knowledge and given our setup, results do not favour the secure use of ROS in industry today, however, we managed to confirm that the security of certain robotic endpoints hold and remain optimistic about securing  ROS industrial deployments.

\end{abstract}

\section{Introduction}


The Robot Operating System (ROS) \cite{quigley2009ros} is the \emph{de facto} framework for robot application development \cite{mayoral2017shift}. 
At the time of writing, the original ROS article \cite{quigley2009ros} has been cited more than 6800 times, which shows its wide acceptance for research and academic purposes. ROS was born in this environment: its primary goal was to provide the software tools that users would need to undertake novel research and development. 
ROS' popularity has continued to grow in industry supported by projects like ROS-Industrial (ROS-I for short)\footnote{https://rosindustrial.org/}, an open-source initiative that extends the advanced capabilities of ROS software to industrial relevant hardware and applications. 

ROS was not designed with security in mind, but as it started being adopted and deployed into products or used in government programs, more attention was placed on it. Some of the early work on securing ROS include \cite{lera2016ciberseguridad, ApplicationSecROS} or \cite{white2016sros}, all of them appearing in the second half of 2016.  At the time of writing, none of these efforts remain actively maintained and the community focus on security efforts has switched to ROS 2, the next generation of ROS. ROS 2 builds on top of DDS \cite{dds} and shows promise. However, to the best of our knowledge, there're still no known robots running ROS 2 in production at scale. From our experience analyzing robots used in industry, their operating systems, libraries and dependencies, we argue that ROS 2 is still years from being widely deployed for major automation tasks. Until then, ROS will prevail.
With the advent of ROS in industry and professional use, one question remains: 
\emph{Even though ROS was not designed with security in mind, can companies use it securely for industrial use cases?} 

The present work tackles this question experimentally. We perform a targeted security exercise, namely \emph{red teaming}, to determine whether ROS and more specifically, ROS and ROS-Industrial packages could be used securely in an industrial setup. We construct a synthetic industrial scenario and choose one of the most common industrial robots with ROS-I support to build it. We then apply available security measures to the setup following official recommendations \cite{stouffer2011guide, IEC62443, canonicalros2020, cisbenchmarkmelodic10} and program a simple flow of operation. 

Using this setup, we perform a red teaming exercise with the overall goal to take control of the ROS computational graph. To achieve this goal, we create four different attacks that target the ROS-Industrial and ROS packages. 
The results show that ROS Melodic Morenia presents several unpatched security flaws, even when hardened with community recommendations.

The remaining content is organized as follows: Section \ref{sec:background} presents related work. Section \ref{sec:use_case} describes the selected industrial application and use case. Section \ref{sec:redteaming} provides a walk-through on the red teaming activity. Finally, Section \ref{sec:conclusionsandfuture} summarizes results and draws some conclusions while hinting on future work actions.

\section{Related work}
\label{sec:background}

Red teaming is a full-scope, holistic and targeted (with specific goals) attack simulation designed to measure how well a system can withstand an attack. Opposed to Penetration Testing (\emph{pentesting} or PT), a red teaming activity does not seek to find as many vulnerabilities as possible to risk-assess them, but has a specific goal. Red teaming looks for vulnerabilities that will maximize damage and meet the selected goals. Its ultimate objective is to test an organization/system detection and response capabilities in production and with respect a given set of objectives. Past works in robot cybersecurity \cite{RobotHaz, lacava2020current, lera2016cybersecurity, clark2017cybersecurity, balsa2017cybersecurity, doczi2016increasing, sandoval2018communications} criticize the current status of cybersecurity in robotics and reckon the need of further research. Previous attempts to review the security of robots via offensive exercises or tools include \cite{mcclean2013preliminary, rctf, olssoniot, mayoral2019industrial, rivera2019rosploit, dieber2020penetration} which mostly focus on proof-of-concept attacks and basic penetration testing, detecting flaws in ROS. A recent study \cite{maggi2020rogue} mentions the identification of several flaws within ROS-Industrial codebase, however it does not explicitly describe  ROS-specific flaws.  Considerations are made with regard the open and insecure architecture predominant in ROS-Industrial deployments throughout its open source drivers. From interactions with the authors of \cite{maggi2020rogue}, it was confirmed that the reported security issues were made generic on purpose, further highlighting the need for further investment on understanding the security landscape of ROS-Industrial setups. 

To the best of our knowledge, no prior public work has performed a red teaming activity on ROS-Industrial packages (or in any other robotics technology for that matter), and challenged its security extensions. In this paper, we present a study in which we aim to do so in a realistic industrial scenario.

\section{Use case}
\label{sec:use_case}

\begin{figure*}[!h]
    \includegraphics[width=1\textwidth]{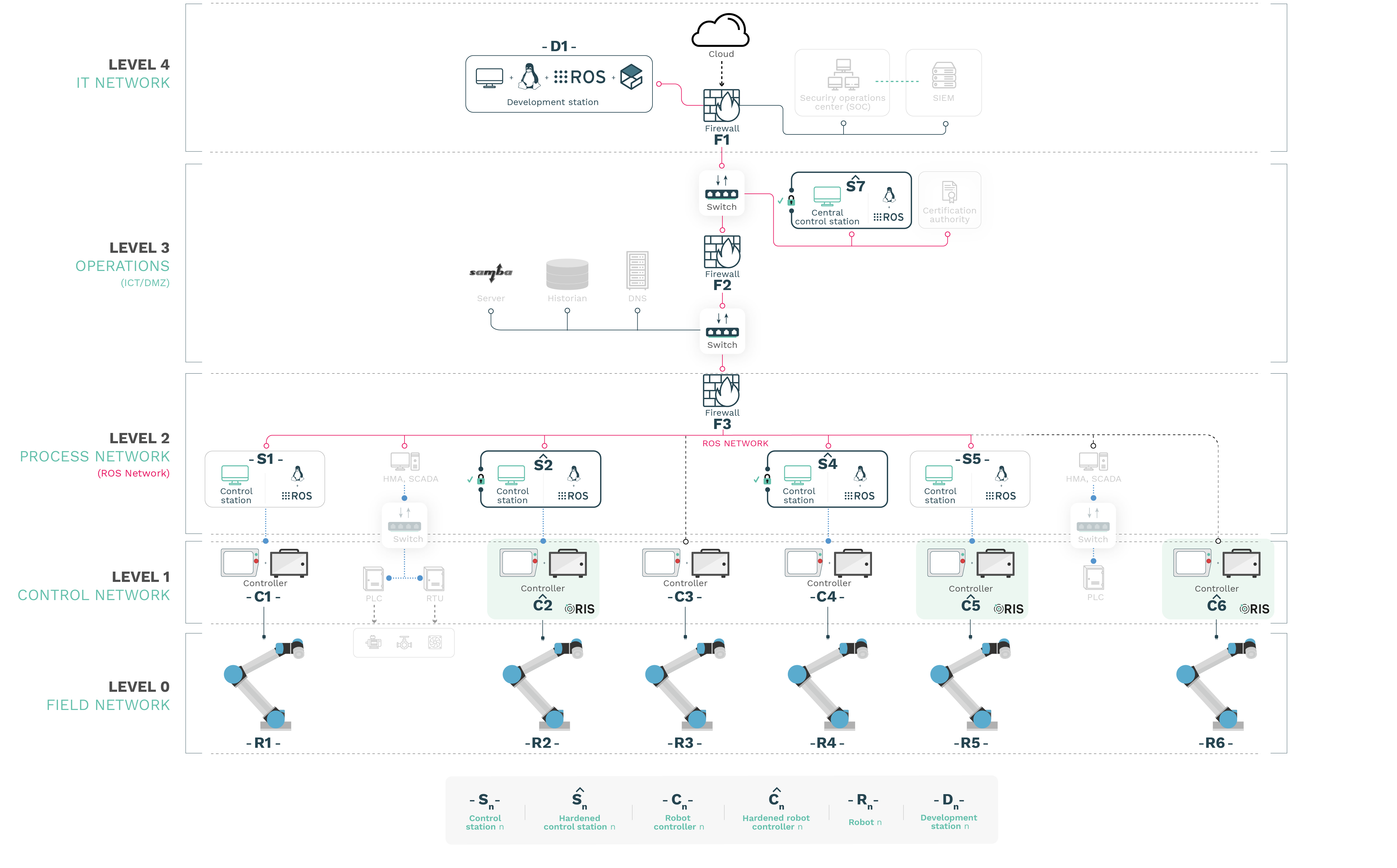}
    \centering
    \caption{
        \footnotesize \textbf{Use case architecture diagram}. The synthetic scenario presents a network segmented in 5 levels with segregation implemented following recommendations in NIST SP 800-82 and IEC 62443 family of standards. There are 6 identical robots from Universal Robots presenting a variety of networking setups and security measures, each connected to their controller. $\hat{S_n}$ and $\hat{C_n}$ denote security hardened versions of an $n$ control station or controller respectively.
    }
    \label{fig:architecture}
\end{figure*}

We build a synthetic assembly line operated by ROS-powered robots while following industrial guidelines on setup and security. The scenario is built following NIST Special Publication 800-82 \cite{stouffer2011guide} Guide to Industrial Control Systems (ICS) Security as well as some parts of ISA/IEC 62443 family of norms \cite{IEC62443}. We segregate the use case in 5 network levels as depicted in Figure \ref{fig:architecture}.

The use case involves several robots with their corresponding controllers. Most of them presented as provided by the manufacturer and some others hardened. For robot (endpoint) hardening we use a commercial Robot Endpoint Protection Platform (REPP) solution applied to the controllers, the Robot Immune System (RIS)\footnote{The Robot Immune System (RIS), \url{https://bit.ly/3gZ9Opu}}. 
In addition, each robot is connected to a Linux-based control station that runs the ROS-Industrial drivers. Control stations are hardened by following the guidelines described in \cite{redteamingrosindustrial_whitepaper}. To simplify, for the majority of the cases we assume that the controller is connected to a dedicated Linux-based control station that runs ROS Melodic Morenia distribution and the corresponding ROS-Industrial driver\footnote{We evaluated both the official ROS-I driver \url{https://bit.ly/2FLaqCl} and the community one \url{https://bit.ly/33273in}}. For those cases that do not follow the previous guideline, the robot controller operates independent from the ROS network (e.g. robots $R_3$ and $R_6$).
To select the target robots, we performed a preliminary evaluation of the different common ROS-Industrial packages. We base our assessment on the potential security bugs identified with static analysis also covered at \cite{redteamingrosindustrial_whitepaper}. 
The results showed that Universal Robots drivers presented the biggest number of bugs which together with its popularity, made us select it as our target. The whole process is documented in \cite{redteamingrosindustrial_whitepaper} which presents an extended version of this study. Figure \ref{fig:architecture} presents the architecture diagram of the use case. To speed up the cybersecurity research and have a common, consistent and easily reproducible development environment, we containerized simulations using \texttt{alurity toolbox}\footnote{\url{https://aliasrobotics.com/alurity.php}}. In most cases, for simulation purposes, the corresponding file systems of each element in the scenario is embed into a Linux container with the right services triggered at launch, facilitating the cooperation across teams of engineers working remotely. The complete use case depicted in Figure \ref{fig:architecture} can be reproduced with the alurity YAML configuration file available for download at \url{https://bit.ly/3lWn41G}.

\section{Red teaming ROS}
\label{sec:redteaming}

We performed a \emph{red teaming} exercise on the ROS network including ROS and ROS-Industrial packages. While targeting ROS, a variety of attack vectors were evaluated. Before diving into the attacks, below, we first specify the goals of the red teaming exercise, defining certain boundaries to  steer our research. After that, we analyze a series of attacks that successfully meet the defined goals.

\subsection{Goals}

For the red teaming exercise, our efforts focus on achieving the following goal:\\

\vspace{-3mm}
\noindent \emph{Goal $G_1$}: Control, deny access or disrupt the ROS computational graph.\\


\vspace{-3mm}
Note that if appropriate security mechanisms are implemented, control of the ROS network might not necessarily imply control of the robots thereby in addition, as a secondary target, we also aim at:\\

\vspace{-3mm}
\noindent \emph{Goal $G_2$}: Control, deny access or disrupt the operation of robots (ROS-powered or not). 

\subsection{Scope}

For the purpose of this \emph{red teaming} exercise and as part of the robotic systems selected, the mechanics are required to be connected to the corresponding controllers, which are the ones operating and interfacing between the robot and other systems. We discard and scope out all activities related to the physical damage of the robot mechanics (servos, encoders and related) by insider threats. All mechanical aspects including malfunctions or related are also considered out of the scope. Similarly and to reduce the complexity of the scenario, while remaining faithful to most industrial deployments, we assume that no wireless connection happens between control stations, robot controllers and/or other devices. In addition, we assume that no social engineering is performed and no kernel exploits are used. 

\subsection{Attack 1 ($A_1$): Targeting ROS-Industrial and ROS core packages}
\label{sec:attack1}

\begin{figure*}[!h]
    \includegraphics[width=1\textwidth]{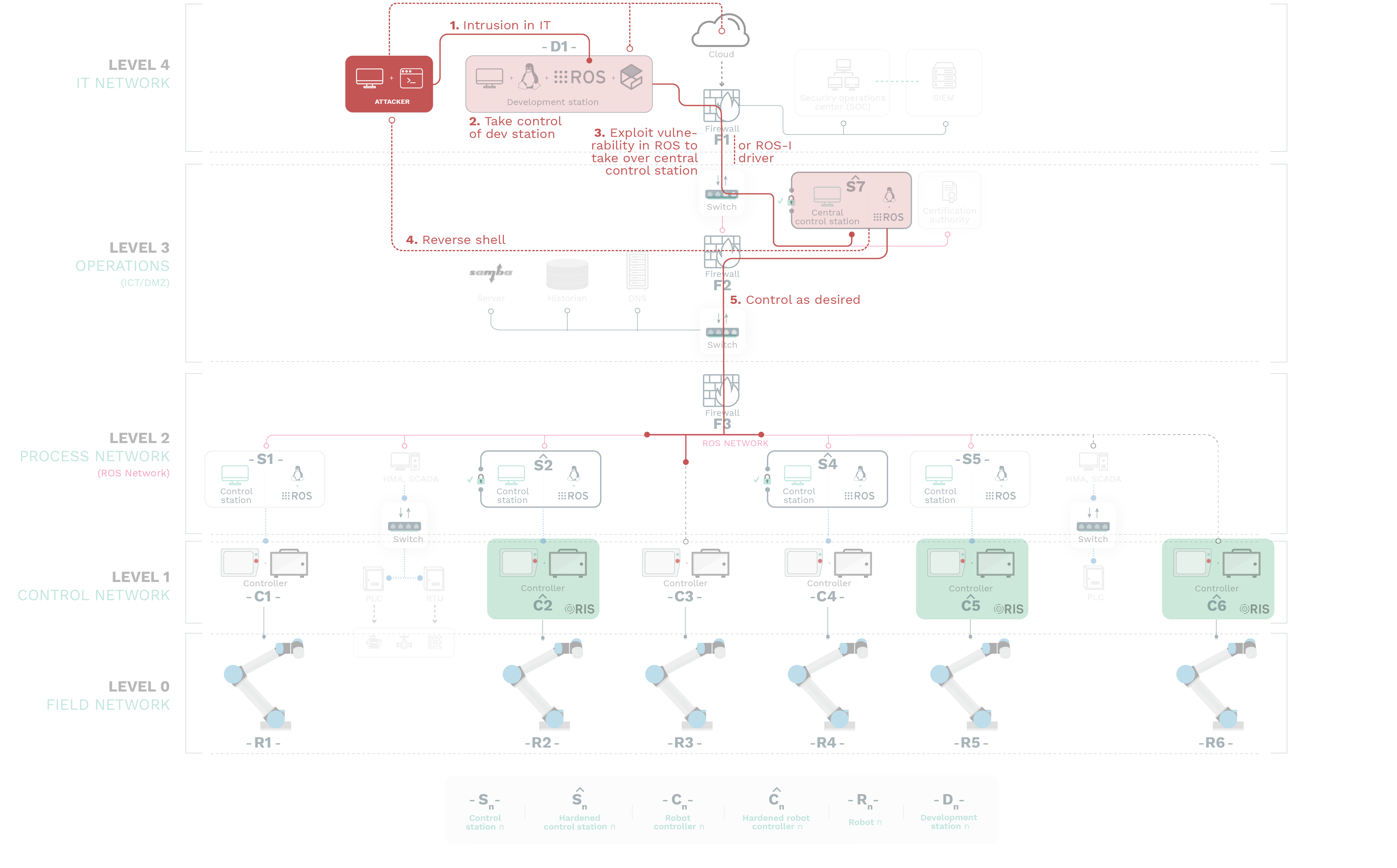}
    \centering
    \caption{
    \footnotesize 
    \textbf{Attack targeting ROS-Industrial and ROS core packages}. The attacker exploits a vulnerability present in a ROS package running on $\hat{S_7}$ (actionlib). Since $\hat{S_7}$ is acting as the ROS Master, segregation does not impose restrictions on it and it is thereby used to access other machines in the OT level to send control commands.
    }
    \label{fig:attack1}
\end{figure*}

In this attack, we adopt the position of an attacker with access to and privileges in a development machine $D_1$ at the IT side (see \emph{Level 4} in Figure \ref{fig:attack1}). Reaching such a machine is beyond the scope of this study but generally consists of an attacker using either a Wide Area Network (WAN) (such as the Internet) or a physical entry-point to exploit an existing vulnerability in the development machine $D_1$ and obtain a certain amount of privileges (\emph{step 1} of the attack diagram of Figure \ref{fig:attack1}). Further, a privilege escalation will be performed by the exploitation of additionally vulnerable services, which allows the attacker to eventually gain privileges into $D_1$ and command it as desired (\emph{step 2}). From $D_1$, an attacker would pivot into \emph{Level 3} by exploiting a vulnerability in the ROS core and/or ROS-Industrial packages (\emph{step 3}). Having gained control of the Central Control Station $\hat{S_7}$ the attacker could decide to establish a reverse channel of communications directly -- avoiding the developer station -- (\emph{step 4}) or proceed to control Operational Technology (OT, \emph{Level 2 and below}) by sending commands via the ROS computational graph (\emph{step 5}). The following subsections detail some of the steps involved on how our team managed to execute steps 3-5.\\

\subsubsection{\emph{Step 3}: exploiting vulnerability in ROS or ROS-Industrial packages for remote code execution}

Since we are targeting $\hat{S_7}$, we scanned the source code of Melodic and the common ROS-Industrial packages being used on it as a ROS Master. We encountered several potentially exploitable flaws and reported them all in RVD \cite{vilches2019introducing}. Then, we decided to focus on one existing flaw in the ROS \texttt{actionlib} package. Part of the ROS core, the \texttt{actionlib} stack provides a standardized interface for interacting with preemptable tasks. Examples of use include exchanging information with an articulated robotic arm (e.g. setting a specific state). In our setup, \texttt{actionlib} is used both by the \texttt{Universal\_Robots\_ROS\_Driver} and the \texttt{ur\_modern\_driver} ROS-Industrial drivers. These are running in the control stations $S_1$, $S_2$, $S_4$ and $S_5$, which interface with robots $R_1$, $R_2$, $R_4$ and $R_5$, respectively. The specific exploitable flaws identified in the \texttt{actionlib} tools are further illustrated in listing \ref{lst:actionlib2}. Note, while this flaw is present in a ROS core package (detected in Melodic, but also in Noetic and prior released ROS distros), the distributed software architecture of ROS propagates this vulnerability to both of the ROS-Industrial drivers mentioned.

\lstset{language=Python}
\lstset{label={lst:actionlib2}}
\lstset{basicstyle=\tiny,
    numbers=left,
    firstnumber=131,
    stepnumber=1,
    commentstyle=\color{lightgray}}
\lstset{caption={
    \footnotesize actionlib \emph{tools/library.py:132}, use of unsafe yaml load vulnerability reported first in RVD\#2401 (\url{https://github.com/aliasrobotics/RVD/issues/2401}). Security flaw highlighted in {\color{redalias} red}.
    } 
}
\lstset{escapeinside={<@}{@>}}
\begin{lstlisting}
def yaml_msgs_str(type_, yaml_str, filename=None):
    import yaml
    <@\textcolor{redalias}{yaml\_doc = yaml.load(yaml\_str)}@>
    msgs = []
    for msg_dict in yaml_doc:
        if not isinstance(msg_dict, dict):
            if filename:
                raise ValueError(yaml file [%s] does not contain a list of dictionaries % filename)
            else:
                raise ValueError(yaml string does not contain a list of dictionaries)
        m = type_()
        roslib.message.fill_message_args(m, msg_dict)
        msgs.append(m)
    return msgs
\end{lstlisting}


\begin{figure}[!h]
    \centering
    \begin{subfigure}[b]{0.45\textwidth}
        \includegraphics[width=0.6\textwidth]{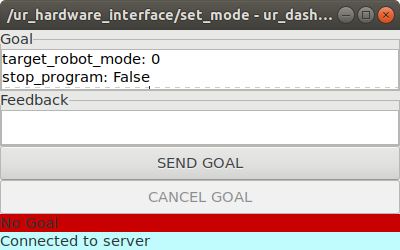}
        \centering
        \caption{\footnotesize  Action client GUI}
        \label{fig:attack1_actionclient}
    \end{subfigure}
    ~ 
    \begin{subfigure}[b]{0.45\textwidth}
        \includegraphics[width=0.9\textwidth]{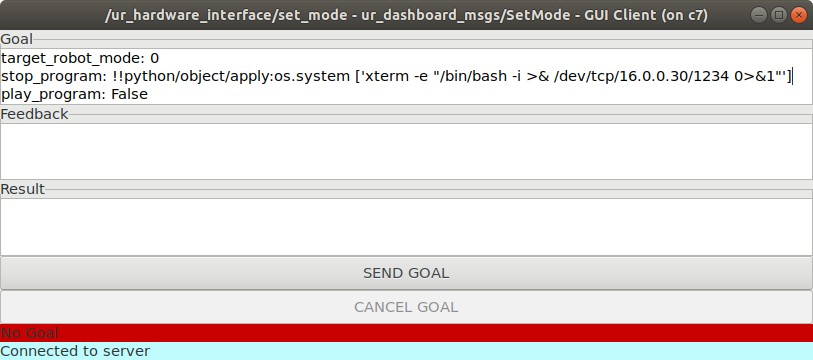}
        \centering
        \caption{\footnotesize  Malicious payload}
        \label{fig:attack1_malicious_payload}
    \end{subfigure}
    \caption{
    \footnotesize 
    \textbf{Remote arbitrary code execution in a machine exploiting a ROS vulnerability with user interaction}. \ref{fig:attack1_actionclient} displays the result of remote launching listing \ref{lst:xml_launchfile} in the attacker's machine ($D_1$) and against the ROS Master target ($\hat{S_7}$). \ref{fig:attack1_malicious_payload} depicts the payload introduced from the attacker's machine ($D_1$) and executed on the target ROS machine ($\hat{S_7}$).
    }
    \label{fig:attack1_screens}
\end{figure}

The flaw itself is caused by an unsafe parsing of YAML values which happens whenever an \textit{action} message is processed to be sent, and allows for the creation of Python objects (\emph{step 3}). 
Through this flaw in the ROS core package of \texttt{actionlib}, an attacker can make $\hat{S_7}$, the central control station that runs ROS Master, execute arbitrary code in Python. Note that \texttt{actionlib} is common in ROS and ROS-Industrial deployments. Also, the selected flaw affects actionlib's tools and depending on the setup, might require certain user interaction for its exploitation. We considered the following two attack scenarios: 

\paragraph{Remote arbitrary code execution: $D_1$ and $\hat{S_7}$ have previously exchanged keys} ($A_{1.1}$). A common (though insecure) practice in industrial environments is to temporarily (or even permanently) exchange keys to facilitate remote control and monitoring of machines in the DMZ level (Level 3). This aligns nicely with the fact that it is common in ROS deployments to rely on SSH key exchanges for remote ROS node launches (via XML launch files\footnote{X11 port forwarding is enabled.}). Correspondingly, we built a custom launch file (listing \ref{lst:xml_launchfile}) that enables us to drop a malicious payload that exploits the vulnerabilities described above. Once a malicious attacker operating from $D_1$ initiates this launch file, it establishes an SSH connection between $D_1$ and $\hat{S_7}$ using pre-shared keys, and forwards the action client GUI visualization to $D_1$ as depicted in Figure \ref{fig:attack1_actionclient}. This way, the attacker can introduce a payload that exploits said vulnerability. We demonstrate this step in Figure \ref{fig:attack1_malicious_payload}. When sent this message causes the action client (\texttt{actionlib}) to execute the malicious payload. The described process allows for arbitrary remote code execution (with the privileges of the ROS setup) exclusively through ROS exploitation. That is, a flaw in ROS allows the attacker to take control of the remote machine $\hat{S_7}$ via common ROS tools.
    
\paragraph{Privilege escalation: Attacker obtains limited access to $\hat{S_7}$ via other means} ($A_{1.2}$). Provided the attacker could execute arbitrary commands on $\hat{S_7}$ for diagnosis (e.g. with a \emph{maintainer} user) but not with a ROS graph privileged one, we believe it's worth further studying whether the exploitation of the same vulnerabilities could lead to obtain privileges that allow to modify the ROS computational graph. Due to time and budget restrictions for the experiment we were not able to confirm this, however we argue that is possible unless ROS specific measures on user privilege-separation have been implemented.

\lstset{
language=XML,
  morekeywords={encoding,
    xs:schema,xs:element,xs:complexType,xs:sequence,xs:attribute, launch, env-loader, node, machine, address, pkg, type, args, env, name}
}
\lstset{label={lst:xml_launchfile}}
\lstset{basicstyle=\tiny,
    numbers=left,
    firstnumber=1,
    stepnumber=1}
\lstset{caption={
        \footnotesize  ROS custom launch file which enables an attacker to deliver a malicious payload in a target ROS machine.
    }
}
\lstset{upquote=true}
\begin{lstlisting}
<launch>
  <env name="DISPLAY" value=":0.0"/>
  <machine name="s7" address="16.0.0.20" env-loader="/opt/ros_ur_ws/devel/env.sh"/>
  <node name="action" machine="s7" pkg="actionlib" type="axclient.py" args="//ur_hardware_interface/set_mode"/>
</launch>
\end{lstlisting}

\subsubsection{\emph{Step 4}: establishing a reverse shell}

With listing \ref{lst:xml_launchfile} remotely executed on the target ROS Master ($\hat{S_7}$) we were able to demonstrate how an attacker can remotely execute arbitrary code. To continue with our attack we seek for a persistent connection and thereby build a custom payload that spawns a reverse shell. The code to do this is depicted in Figure \ref{fig:attack1_malicious_payload}. It constructs a string which when processed for generating ROS communication artifacts (messages), gets executed. The string itself declares a Python object which on creation launches a reverse shell back to the attacker's ($D_1$) hardcoded IP address.

\subsubsection{\emph{Step 5}: control the computational graph and other machines within the OT levels}

Once the attacker has at its disposal a reverse shell to $\hat{S_7}$  it becomes relatively easy to command the different industrial subsystems. $\hat{S_7}$ acts as the ROS Master of the industrial network and thereby can easily influence all ROS-Industrial package deployments living in the control stations $S_1$ to $S_5$. Such exploitation has been covered by other authors including \cite{dieber2020penetration} and we refer the reader to this study for further exploration on how to take control of the ROS computational graph using the ROS Master and Slave APIs.

\subsubsection{Responsible disclosure and mitigation efforts}
Our team announced the Robot Vulnerability Database in October 2019 for the ROS community\footnote{\url{https://bit.ly/2GBgk9v}} and openly disclosed our intention of cataloging and recording early-phase security flaws applying to ROS. The flaw described in here was first publicly filed in June 2020 and later elevated to a vulnerability in August 2020 with subsequent pull requests patching \texttt{actionlib} in ROS Melodic Morenia\footnote{\url{https://github.com/ros/actionlib/pull/170}} and ROS Noetic Ninjemys\footnote{\url{https://github.com/ros/actionlib/pull/171}}. The suggested mitigations propose the use of safe parsing. This way, the construction of communication artifacts would only allow for simple objects like strings or integers, removing the threat.

\subsection{\emph{Attack 2 ($A_2$)}: Disrupting ROS-Industrial communications by attacking underlying network protocols}

As pointed out previously, ROS-Industrial software builds on top of ROS packages which build on top of traditional networking protocols at OSI layers 4 and 3. It is common to find ROS deployments using TCP/IP in the Transport and Network levels of the communication stack. To further test the limits of these underlying layers and its impact on ROS, we developed a complete ROSTCP networking package dissector and used it as a tool for attacks. These consists of a malicious attacker with privileged access to an internal ROS-enabled control station (e.g. $S_1$) disrupting the ROS-Industrial communications and interactions between other participants of the network. The attack leverages the lack of authentication in the ROS computational graph previously reported in other vulnerabilities of ROS such as \href{https://github.com/aliasrobotics/RVD/issues/87}{RVD\#87} or \href{https://github.com/aliasrobotics/RVD/issues/88}{RVD\#88}.

By simply spoofing another participant's credentials (at the Network level) and either disturbing or flooding communications within infrastructure's \emph{Level 2} (Process Network), we are able to heavily impact the ROS and ROS-Industrial operation\footnote{The execution of these attacks required us to develop a package dissector/crafter and configure the attacker's kernel to ignore certain types of network requests so that it does not conflict with the attacking activity. Details on this have been purposely omitted.}. Our team considered 
a FIN-ACK attack which aims to disrupt network activity by saturating bandwidth and resources on stateful interactions (i.e. TCPROS sockets). 
The simple proof-of-concept we developed for validating this flaw can be downloaded from \url{https://bit.ly/3h5Fn11}. Further details are available in our extended report \cite{redteamingrosindustrial_whitepaper}.
\subsection{Attack 3 ($A_3$): Person-In-The-Middle attack to a ROS control station}

A Person-in-the-Middle (PitM) attack targeting a control station (e.g. $\hat{S_2}$) consists of an adversary gaining access to the network flow of information and sitting in the middle, interfering with communications between the original publisher and subscriber as desired. PitM demands to conflict not just with the resolution and addressing mechanisms but also to hijack the control protocol being manipulated (ROSTCP). The attack gets initiated by a malicious actor gaining access and control of a machine in the network (\emph{step 1}), such as done with \emph{$A_1$}. Then, using the compromised computer on the control network, the attacker poisons the ARP tables on the target host ($\hat{S_7}$) and informs its target that it must route all its traffic through a specific IP and hardware address (\emph{step 2}), i.e., the attacker’s owned machine. By manipulating the ARP tables, the attacker can insert themselves between $\hat{S_7}$ and $\hat{S_2}$ (\emph{step 3})\footnote{The attack described in here is a specific PitM variant known as ARP PitM.}. When a successful PitM attack is performed, the hosts on each side of the attack are unaware that their network data is taking a different route through the adversary’s computer. 

Once adversaries have successfully inserted their machine into the information stream, they have full control over the  communication and could carry out several types of attacks. For example, the replay attack (\emph{step 4}). In its simplest form, captured data from $\hat{S_7}$ is replayed or modified and replayed. During this replay attack the adversary could continue to send commands to the controller and/or field devices to cause an undesirable event while the operator is unaware of the true state of the system. 





\subsection{Attack 4 ($A_4$): Exploiting known vulnerabilities in a robot endpoint to compromise the ROS network}

Attacks do not only necessarily come from the outside (IT Level or the Cloud). Increasingly more reports \cite{filkins2019sans} are informing about the relevance of insider threats with more than half of the attack vectors requiring physical access. We studied one of such scenarios where we attempted first to compromise $\hat{C_6}$ (failed) and then $C_3$ using previously reported and known (yet unresolved) zero day vulnerabilities in the robot controller\footnote{Examples of past zero day attacks include \href{https://github.com/aliasrobotics/RVD/issues/1413}{RVD\#1413 }, \href{https://github.com/aliasrobotics/RVD/issues/1410}{RVD\#1410},  \href{https://github.com/aliasrobotics/RVD/issues/673}{RVD\#673} or \href{https://github.com/aliasrobotics/RVD/issues/1408}{RVD\#1408} among others.}. Due to the lack of concerns for security from manufacturers, these end-points can easily become rogue and serve as an entry point for malicious actors. After failing to take over the hardened control station, our team successfully prototyped a simplified attack using \href{https://github.com/aliasrobotics/RVD/issues/1495}{RVD\#1495} (CVE-2020-10290) and taking control over $C_3$. From that point on, we could access the ROS network completely and pivot (\emph{$A_1$}), disrupt (\emph{$A_2$}) or PitM (\emph{$A_3$}) as desired.


\section{Discussion}
In this section, we summarize and discuss our findings and lessons learned.

\subsection{Findings}
Table \ref{table:incidents} summarizes the attacks and their impact with respect to our two goals. $G_1$ is achieved in all the presented attacks whereas $G_2$ is mostly achieved yet depends on the hardening of the corresponding control stations and robotic endpoints. 
At the time of writing, among the vulnerabilities we exploited most remain active. An exception is \href{https://github.com/aliasrobotics/RVD/issues/2401}{RVD\#2401} which got resolved by Open Robotics within 30 hours from the moment we submitted a mitigation.

\setlength{\tabcolsep}{10pt}
\renewcommand{\arraystretch}{1.5}
\begin{table}[h!]
    \centering
    \caption{
    \footnotesize 
        Summarizes security incidents demonstrated for the elected industrial use case as part of the red teaming exercise. $R_n$ refers to the $n$ robot of the use case as depicted in Figure \ref{fig:architecture}.
    }
    \scalebox{0.65}{
    \begin{tabular}{ |p{2.2cm}|p{5.9cm} | p{1.2cm} | } 
        \hline
        \textbf{Attack} & \textbf{Description} & \textbf{Goals met} \\
        \hline

        
        \emph{$A_{1.1}$}: remove arbitrary code execution & Subject to some prior interactions, attacker with control of $D_1$ is able to exploit a vulnerability in ROS and launch arbitrary remote code executions from a privileged ROS end-point compromising completely the computational graph & $G_1$ and $G_2$ ($R_1$, $R_2$, $R_3$, $R_4$ and  $R_5$) \\ 
        \hline
        
        \emph{$A_{1.2}$}: privilege escalation & Subject to local access, attacker is able to exploit a vulnerability in ROS and escalate privileges (to the ROS ones) in such machine & $G_1$ \\ 
        \hline


        \emph{$A_{2}$}: FIN-ACK flood attack targeting ROS & Attacker attempts to deny ROSTCP connection on target destination by forcing a maxed-out number of connections & $G_1$ and $G_2$ ($R_1$, $R_2$, $R_3$, $R_4$ and $R_5$) \\ 
        \hline

        \emph{$A_3$}: PitM attack to a ROS control station & Attacker poisons ARP tables and gains access to the network flow of information siting between targeted publishers and subscribers, interfering with communications as desired.  & $G_1$ and $G_2$ ($R_1$, $R_2$, $R_3$, $R_4$ and $R_5$) \\ 
        \hline

        \emph{$A_4$}: Insider endpoint via unprotected robot controller & Attackers exploit known vulnerabilities in a robot endpoint to compromise the controller and pivot into the ROS network. & $G_1$ and $G_2$ ($R_1$, $R_2$, $R_3$, $R_4$, $R_5$ and $R_6$) \\
        \hline

    \end{tabular}}
    \label{table:incidents}
\end{table}

The original research question posed whether ROS could be used securely on industrial use cases. Based on our experimental results, we found: \emph{With the current status of ROS, it is hardly possible to guarantee security without additional measures.} 

\subsection{Lessons learned}

\noindent Through our experiments we showed how control stations running Ubuntu 18.04 do not protect ROS or ROS-Industrial deployments. Moreover, the guidelines offered by Canonical \cite{canonicalros2020} for securing ROS were of little use against targeted attacks. Certain ongoing hardening efforts for ROS Melodic \cite{cisbenchmarkmelodic10} helped mitigate some issues but as highlighted in Table \ref{table:incidents}, most goals were still achieved with attacks targeting threats like zero day vulnerabilities, wide and availability of industrial components, inadequate security practices or non-patched OS and firmware.

Dedicated robotic security protection systems like the Robot Immune System (RIS) \cite{robotimmunesystem} used in $\hat{C_2}$, $\hat{C_5}$ or $\hat{C_6}$ managed to secure the corresponding robot avoiding directed attacks. However $R_2$ and $R_5$ robots were still \emph{hijacked} by compromising the ROS computational graph via their control stations. RIS was not able to stop these attacks because they came from trusted sources whose behavior was learned over a prior training phase. An exception was $R_6$ which we were not able to compromise thanks RIS being installed at $\hat{C_6}$ whereas $R_3$ (not protected) was easily compromised and used as a rogue endpoint for attackers to pivot into other malicious endeavors. From this, we conclude that industrial scenarios like the one presented in this use case using ROS must not only follow ICS guidelines \cite{stouffer2011guide, IEC62443} but also harden robot endpoints and the ROS computational graph across each phase, from development to post-production \cite{mayoral2020devsecops}.

\section{Conclusions}
\label{sec:conclusionsandfuture}

In this study, we presented four targeted attacks over a synthetic industrial scenario constructed by following international ICS cybersecurity standards where the control logic is operated by ROS and ROS-Industrial packages. Our attacks exploited both new and known vulnerabilities of ROS achieving the two goals we set. We managed to execute code remotely (\emph{$A_1$}) in a ROS end-point, disrupt the ROS computational graph (\emph{$A_2$}), impersonate a ROS control station through PitM (\emph{$A_3$}) and finally use an unprotected robot endpoint to pivot into the ROS network (\emph{$A_4$}). 
For future work we plan to look into other Operating Systems as a starting point for secure ROS deployment and explore additional security measures.

\section*{ACKNOWLEDGMENTS}

\noindent This research has been partially funded by Alias Robotics and by the European Union‘s Horizon 2020 research and innovation programme under grant agreement No 732287, under ROSin project through the FTP RedROS-I.  Thanks also to the Basque Government, throughout the Business Development Agency of the Basque Country (SPRI). Special thanks to BIC Araba and the Basque Cybersecurity Centre (BCSC) for the support provided. Finally, this research was also financially supported by the Spanish Government through CDTI Neotec actions (SNEO-20181238). \\

\bibliographystyle{IEEEtran}
\bibliography{IEEEabrv,bibliography}

\end{document}